\documentclass{article}

 \usepackage[dblblindworkshop, final]{neurips_2025}
\workshoptitle{Learning from Time Series for Health}
\usepackage[utf8]{inputenc} % allow utf-8 input
\usepackage[T1]{fontenc}    % use 8-bit T1 fonts
\usepackage{hyperref}       % hyperlinks
\usepackage{url}            % simple URL typesetting
\usepackage{booktabs}       % professional-quality tables
\usepackage{amsfonts}       % blackboard math symbols
\usepackage{nicefrac}       % compact symbols for 1/2, etc.
\usepackage{microtype}      % microtypography
\usepackage{xcolor}         % colors

\usepackage{graphicx}
\usepackage{pifont} % checkmark and x symbols

\usepackage{booktabs} % they suggested for tables

\title{Transferring Clinical Knowledge into ECGs Representation}
\author{%
  Jose Geraldo Fernandes \\
  Department of Computer Science\\
  Universidade Federal de Minas Gerais\\
  \texttt{josegeraldof@ufmg.br} \\
  \And
  Luiz Facury de Souza \\
  Department of Computer Science \\
  Universidade Federal de Minas Gerais \\
  \texttt{luizfysouza@ufmg.br} \\
  \AND
  Pedro Robles Dutenhefner \\
  Department of Computer Science \\
  Universidade Federal de Minas Gerais \\
  \texttt{pedroroblesduten@ufmg.br} \\
  \And
  Gisele L. Pappa  \\
  Depart. of Math. and Comp. Sc. \\
  Albion College \\
  \texttt{glpappa@dcc.ufmg.br} \\
  \And
  Wagner Meira Jr.  \\
  Department of Computer Science \\
  Universidade Federal de Minas Gerais \\
  \texttt{meira@dcc.ufmg.br} \\
}

\begin{document}

\maketitle

\begin{abstract}
  Deep learning models have shown high accuracy in classifying electrocardiograms (ECGs), but their black box nature hinders clinical adoption due to a lack of trust and interpretability. To address this, we propose a novel three-stage training paradigm that transfers knowledge from multimodal clinical data (laboratory exams, vitals, biometrics) into a powerful, yet unimodal, ECG encoder. We employ a self-supervised, joint-embedding pre-training stage to create an ECG representation that is enriched with contextual clinical information, while only requiring the ECG signal at inference time. Furthermore, as an indirect way to explain the model's output we train it to also predict associated laboratory abnormalities directly from the ECG embedding. Evaluated on the MIMIC-IV-ECG dataset, our model outperforms a standard signal-only baseline in multi-label diagnosis classification and successfully bridges a substantial portion of the performance gap to a fully multimodal model that requires all data at inference. Our work demonstrates a practical and effective method for creating more accurate and trustworthy ECG classification models. By converting abstract predictions into physiologically grounded \emph{explanations}, our approach offers a promising path toward the safer integration of AI into clinical workflows.
\end{abstract}

\section{Introduction}

% brief introduction of ecg exams, their importance (cardiovascular diseases remain the leading cause of global mortality \cite{roth2017global}) and automatic classification with deep learning
Electrocardiogram (ECG) exams are a fundamental diagnostic tool in medical practice, recording the heart's electrical activity to detect a wide range of cardiovascular conditions \citep{braunwald2015braunwald}. Their importance is substantial, especially considering that cardiovascular diseases remain the leading cause of global mortality \citep{roth2020global}. With the advancement of artificial intelligence, the automatic classification of ECG exams using deep learning techniques has shown remarkable potential \citep{liu2021deep,petmezas2022state}, offering high precision in identifying abnormalities and extracting complex information that can aid in diagnosis.

% then talk about how practitioners don't trust/use healthcare deep learning models, despite their significantly performance, for lack of explainability
Despite the promising performance and significant accuracy achieved by deep learning models in healthcare, their adoption and trust among healthcare professionals remain a challenge \citep{reyes2020interpretability,adeniran2024explainable}. The main barrier to integrating these models into clinical practice lies in their black box nature—the lack of explainability and transparency about how decisions are made \citep{rosenbacke2024explainable}. Physicians, accustomed to clinical reasoning based on evidence and causality, hesitate to trust systems whose internal processes are opaque, leading to mistrust and uncertainty regarding the safety and justification of the predictions \citep{koccak2025bias}.

% Frequently, the explanations provided by deep learning models are presented as saliency maps, which highlight regions in the input space most relevant to the model's prediction \citep{chaddad2023survey}. However, these heatmaps often fail to bridge the trust gap. Their limitations include a lack of robustness and a frequent misalignment with clinically relevant concepts, making their interpretation ambiguous for diagnostic reasoning \citep{borys2023explainable,zhang2023revisiting}. In contrast, explanations that are grounded in the established clinical lexicon—such as concrete physiological abnormalities—are far more valuable, as they align with the workflow of professionals who rely on structured data like laboratory results to make decisions \citep{hulsen2023explainable,gupta2024comparative}.

% To address this explainability gap, we propose a novel multimodal training architecture that enriches an ECG model with knowledge from associated tabular clinical data. Instead of requiring all data modalities at inference, our approach uses a self-supervised, joint-embedding objective to transfer the rich context from laboratory values, vitals, and biometrics into a powerful, unimodal ECG encoder. This enriched encoder is not only more accurate but also inherently more interpretable. By training it to perform a secondary task of predicting lab abnormalities from the ECG signal alone, we create a system that can explain its diagnostic reasoning in terms of concrete, clinically relevant concepts.
Frequently, explanations are presented as saliency maps, which often fail to bridge the trust gap due to a lack of robustness and misalignment with clinical concepts \citep{borys2023explainable, zhang2023revisiting}. To address this, we propose a novel multimodal training architecture that enriches an ECG model with knowledge from associated tabular clinical data. Instead of requiring all data modalities at inference, our approach uses a self-supervised, joint-embedding objective to transfer the rich context from laboratory values and vitals into a powerful, unimodal ECG encoder. By training it to perform a secondary task of predicting lab abnormalities from the ECG signal alone, we create a system that can explain its diagnostic reasoning in terms of concrete, clinically relevant concepts.

In summary, we make the following contribution:
\begin{itemize}
    % \item joint-embedding to distill other-mode knowledge in ecg classification
    \item We propose a joint-embedding pre-training framework to transfer knowledge from multimodal tabular data into a unimodal ECG encoder for the task of diagnosis classification;
    % \item reconstruction of other modes as a alternative to explain automatic diagnoses from ecg
    \item We introduce the prediction of laboratory abnormalities from the ECG embedding as a novel, clinically-grounded method for a indirect explanation of the model's diagnostic outputs.
\end{itemize}

\paragraph{Related Work}
Our work integrates insights from three key areas. While deep learning for ECG classification is well-established, models often lack the trust of clinicians due to their "black box" nature \citep{reyes2020interpretability}. Current eXplainable AI (XAI) methods often rely on saliency maps, which can be misaligned with clinical reasoning \citep{borys2023explainable}. Concurrently, self-supervised learning (SSL) has enabled the creation of powerful representations without labeled data \citep{zbontar2021barlow}, and multimodal learning seeks to create holistic models by fusing data sources \citep{kline2022multimodal}. However, most multimodal approaches require all data at inference time, hindering practicality \citep{alcaraz2025enhancing}. We are the first to unify these areas, using a multimodal SSL objective to distill knowledge into a practical, unimodal-at-inference model with a novel, clinically-grounded explanation mechanism. A detailed literature review is provided in the Appendix.

\section{Method}
\begin{figure}[ht]
  \centering
  \centerline{\includegraphics[width=\columnwidth]{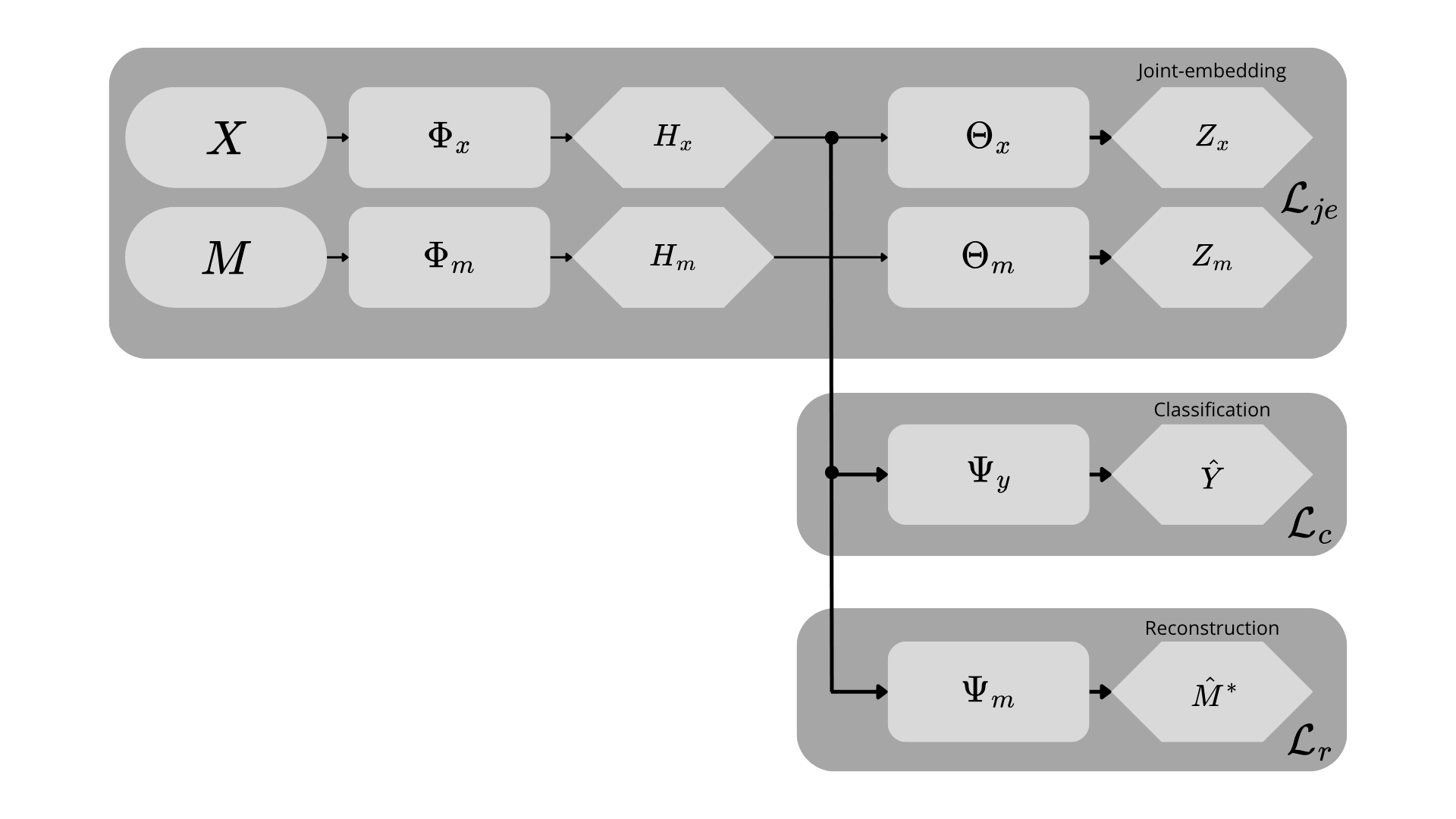}}
  \caption{Schematic of the proposed multimodal training architecture. Our method begins with a (i) joint-embedding pre-training stage, where an ECG encoder $\Phi_x$ learns to produce representations $H_x$ that are aligned, task $\mathcal{L}_{je}$, with embeddings from tabular clinical data $M$. Subsequently, this single, enriched encoder is finetuned for two downstream tasks: (ii) a primary multi-label diagnosis classification task $\mathcal{L}_{c}$; and, (iii) a secondary laboratory abnormality reconstruction-like task $\mathcal{L}_{r}$, which provides a mechanism for aiding decision making. Crucially, only the ECG signal $X$ is required at inference time.}
  \label{fig:architecture}
\end{figure}

Our approach is a three-stage training paradigm designed to create a powerful ECG representation aligned with clinically relevant outputs. We first pre-train a waveform encoder using a self-supervised, cross-modal objective to transfer knowledge from tabular clinical data. Subsequently, we fine-tune this encoder for two downstream relevant clinical tasks: a primary task of multi-label diagnosis classification \cite{strodthoff2024prospects}; and, a secondary task of laboratory values abnormality prediction \cite{alcaraz2024cardiolab}. We organised this algorithm into a fluxogram, as despicted in Figure~\ref{fig:architecture}.

\subsection{Joint-embedding Pre-training}
% should describe the barlow twins setup, using the joint-embedding loss to distill the information from the tabular data into the signal encoder

% we encode both the signal $x$ and tabular data $m$ with encoders $\Phi_x$ and $\Phi_m$ to $h_x$ and $h_m$, then align a projection of their embeddings $z_x$ and $z_m$ with projectors $\Theta_x$ and $\Theta_m$ with the barlow twins loss function $J_{je}$, like equation \ref{eq:je}

The primary goal of our pre-training stage is to learn a signal encoder, $\Phi_x$, whose representations are enriched with the contextual information present in associated tabular clinical data. To achieve this, we frame the task as a self-supervised, joint-embedding problem where the model learns to align representations from these two distinct modalities.

For a given patient encounter, we have a raw ECG signal segment $x \in \mathbb{R}^{C \times L}$, where $C$ is the number of leads and $L$ is the sequence length, and a corresponding vector of tabular clinical data $m \in \mathbb{R}^{D}$, which includes demographics, vitals, and biometrics. The signal and tabular data are processed by their respective encoders, a powerful S4-based sequence model $\Phi_x$ \citep{strodthoff2024prospects} and a MLP $\Phi_m$. These backbones produce feature representations $h_x = \Phi_x(x)$ and $h_m = \Phi_m(m)$. Following the standard practice in self-supervised learning, these feature representations are then mapped into another embedding space using dedicated projector networks, $\Theta_x$ and $\Theta_m$, $z_x = \Theta_x(h_x)$ and $z_m = \Theta_m(h_m)$, $z_x, z_m \in \mathbb{R}^{E}$, where $E$ is the embedding dimension.

To align these embeddings without risking representational collapse, we employ the Barlow Twins loss function, $\mathcal{L}_{BT}$. This objective function does not rely on negative sampling and instead encourages the cross-correlation matrix computed between the embeddings $Z_x$ and $Z_m$ over a batch of $N$ samples to be close to the identity matrix. The loss is composed of two terms: an invariance term that pulls the embeddings from the same patient together; and, a redundancy reduction term that decorrelates the different dimensions of the embedding vectors. The total pre-training objective is thus defined as in Equation~\ref{eq:je}.

\begin{equation}
\label{eq:je}
\mathcal{L}_{je} = \mathcal{L}_{BT}(Z_x, Z_m) = \sum_{i} (1 - C_{ii})^2 + \lambda \sum_{i} \sum_{j \neq i} C_{ij}^2
\end{equation}

where $C$ is the cross-correlation matrix computed between the batch-normalized embeddings $Z_x$ and $Z_m$, and $\lambda$ is a hyperparameter balancing the two terms. By minimizing this loss, the signal encoder $\Phi_x$ is trained to produce representations $h_x$ that are not only descriptive of the ECG signal itself but are also highly predictive of the rich, contextual information contained in the clinical data $m$. This process effectively transfers the clinical knowledge into the parameters of the waveform encoder.

\subsection{Diagnosis Classification}
% should describe the standard classification approach, i think we dont need to spend much space in this subsection

% we then finetune the signal encoder $\Phi_x$ to predict the diagnoses labels $y$ with a predictor $\Psi_y$ and train with bce loss function $\mathcal{L}_{BCE}$

Following the self-supervised pre-training stage, we leverage the learned, context-aware signal encoder $\Phi_x$ for the primary downstream task of clinical diagnosis prediction. To accomplish this, we adopt a standard fine-tuning protocol.

The pre-trained encoder $\Phi_x$ is partially frozen to prevent catastrophic forgetting of the transferred clinical knowledge. A new, task-specific classification head, $\Psi_y$, is then initialized and attached to the encoder. This head is a lightweight MLP that takes the feature representation $h_x = \Phi_x(x)$ and maps it to a logit for each of the $K$ possible diagnoses.

The model is then trained on the subset of the data containing ground-truth diagnosis labels $Y$. The optimization objective is the standard Binary Cross-Entropy (BCE) loss, summed over all possible labels, as is common for multi-label classification problems. The classification loss is defined as in Equation~\ref{eq:bce}.

\begin{equation}
\label{eq:bce}
\mathcal{L}_{c} = \mathcal{L}_{BCE}(Y, \hat{Y}) = -K^{-1} \sum_k^K [Y_k \log(\hat{Y}_k) + (1 - Y_k) \log(1 - \hat{Y}_k)]
\end{equation}

where $Y \in \{0, 1\}^K$ is the one-hot encoded vector of true labels and $\hat{Y}$ is the vector of predicted logits from the model. This fine-tuning stage adapts the powerful, general-purpose representation learned during pre-training to the specific patterns required for the diagnostic task.

\subsection{Reconstruction Finetuning}
% should describe the estimation task, predicting abnormalities of some features of the tabular data (laboratory exam related)

% we then use this signal encoder $\Phi_x$ for the reconstruction task, we want to use this prediction $\hat{m}^*$ as a way to explain the classification task, note however that we are not completed reconstructing the data $m$ but instead predicting a logit to identify abnormalities in some lab exam features (estimation task), in this context $m^*$ represents the one-hot encoding space of the abnormalities. we use another predictor $\Psi_m$ and train with bce loss function $\mathcal{L}_{BCE}$

A primary motivation for our three-stage approach is to produce a waveform representation that is not only predictive but also more interpretable. We train the signal encoder $\Phi_x$ to perform a reconstruction task of laboratory abnormalities that are related to the context seem in the pre-training.

For this purpose, we introduce a second, independent head $\Psi_m$, which is attached to the same pre-trained encoder $\Phi_x$. This head is trained to predict a multi-label vector $M^* \in \{0, 1\}^P$, where $P$ is the number of distinct laboratory abnormalities defined in our dataset (e.g., "Hemoglobin\_high", "Urea Nitrogen\_low"). Each element $m^*_p \in M^*$ is a binary indicator of a specific lab abnormality.

Similar to the classification phase, the reconstruction head $\Psi_m$ is a lightweight MLP that takes the signal representation $h_x$ and outputs a vector of logits $\hat{M}^* = \Psi_m(\Phi_x(x))$, one for each of the $P$ lab abnormalities. The model is trained by minimizing the BCE loss, as defined in Equation~\ref{eq:bce}, now applied to the lab abnormality targets $m^*$ as in Equation~\ref{eq:recon}.

\begin{equation}
\label{eq:recon}
\mathcal{L}_{r} = \mathcal{L}_{BCE}(M^*, \hat{M}^*)
\end{equation}

Our hypothesis is that the ability of the model to successfully perform this pseudo-reconstruction task, using only the ECG signal as input, serves to provide a mechanism for model explainability. For any given diagnosis prediction, we can now also output the concurrent lab abnormalities predicted from the same ECG embedding, offering clinicians a direct, data-driven insight into the potential physiological drivers behind the model's primary prediction. Essentially we offer something like: "The model predicts diagnosis $Y$, and it's doing so as it also is predicting a associated lab abnormality $M^*_p$ (e.g., high creatinine)."

\section{Experimental Setup and Results}
We evaluate our paradigm on the MIMIC-IV-ECG dataset \citep{johnson2023mimic}, focusing on the emergency department subset per the protocol in \citet{alcaraz2025enhancing}, which also defines our data splits and backbone architecture, we detail the experiment setup in the Appendix. We report macro-averaged AUROC and compare against three key baselines: a \textbf{Supervised Signal-Only} model \citep{strodthoff2024prospects}; a \textbf{Multimodal Lab Prediction} model serving as an upper bound for the reconstruction task \citep{alcaraz2024cardiolab}; and a fully \textbf{Multimodal Classification} model serving as a practical upper bound for diagnosis \citep{alcaraz2025enhancing}.

\begin{table*}[t]
\caption{Main results comparing our joint-embedding (JE) and reconstruction approach to established baselines. Filled circles ($\bullet$) indicate a data requirement, while empty circles ($\circ$) indicate it is not required.}
\label{tab:results}
\centering
% Using \small to make the table a bit smaller without resizing the whole box
\resizebox{\textwidth}{!}{ % idk if they enforce the font size
\begin{tabular}{@{}l cc ccc ccc@{}}
\toprule
& \multicolumn{2}{c}{\textbf{Evaluation}} & \multicolumn{3}{c}{\textbf{Inference Requirement}} & \multicolumn{3}{c}{\textbf{Training Requirement}} \\
\cmidrule(lr){2-3} \cmidrule(lr){4-6} \cmidrule(lr){7-9}
\textbf{Model} & \textbf{Diagnoses} & \textbf{Lab} & \textbf{Routine} & \textbf{Lab} & \textbf{Diagnoses} & \textbf{Routine} & \textbf{Lab} & \textbf{Diagnoses} \\
\midrule
Supervised Signal-Only & 0.768 & - & $\circ$ & $\circ$ & $\circ$ & $\circ$ & $\circ$ & $\bullet$ \\
Multimodal Lab Prediction & - & \textbf{0.762} & $\bullet$ & $\circ$ & $\circ$ & $\bullet$ & $\bullet$ & $\circ$ \\
Multimodal Classification & \textbf{0.826} & - & $\bullet$ & $\bullet$ & $\circ$ & $\bullet$ & $\bullet$ & $\bullet$ \\
\midrule
\textbf{Ours} & 0.795 & 0.701 & \textbf{$\circ$} & \textbf{$\circ$} & \textbf{$\circ$} & \textbf{$\bullet$} & \textbf{$\bullet$} & \textbf{$\bullet$} \\
\bottomrule
\end{tabular}
}
\end{table*}

As shown in Table~\ref{tab:results}, the results suggest the effectiveness of our approach. Our final model, which operates using only the ECG signal at inference time, achieves a classification performance improvement over the signal-only baseline indicating the presence of tabular data knowledge transfer. While it does not reach the performance of the multimodal upper bound which requires full access to all data at test time, our method successfully bridges a significant portion of this performance gap without sacrificing the practicality of unimodal deployment.

Furthermore, we also assess the indirect explanations by evaluating its performance on the laboratory abnormality prediction task. Our approach achieves similar performance to its correspondence baseline, again we treat them as an upper bound because of the data requirement at inference. 

\section{Limitations and Conclusion}
% should discuss the limitations and conclude

% \paragraph{three stage sequential training} this is arguably the main limitation, because of the sequential learning the model could 'forget' the knowledge from the previous training stage. we've tried to train jointly but it was very hard to optimize the three losses. we try to prevent this by partially freezing the earlier layers of the signal encoder. a regularization with label from the previous training might as well be a good solution.
\paragraph{Sequential Training Paradigm} A primary limitation of our work lies in the three-stage sequential training paradigm. This approach introduces the risk of catastrophic forgetting \citep{mccloskey1989catastrophic}, where the model may lose some of the rich, transferred knowledge from the pre-training stage during subsequent finetuning. While a joint, multi-task learning approach was considered, we found the simultaneous optimization of the three distinct loss functions to be highly challenging and unstable. To mitigate this, we partially froze the early layers of the signal encoder, preserving the foundational representations learned during pre-training. Future work could explore more advanced continual learning strategies to more robustly preserve the pre-trained knowledge. Promising directions include (i) regularization-based methods, such as using the pre-trained embeddings as a regularization target during finetuning, or (ii) representation distillation, where a separate "student" ECG encoder is explicitly trained to mimic the output representations of a powerful, pre-trained multimodal "teacher" model.

% \paragraph{explanation evaluation} we evaluate our explanation with the performance on the estimation task, however this is assessment is indirect. we didn't evaluate the causal relationships between the lab abnormalities and diagnoses, neither indicate which of them are related to the practitioner.
\paragraph{Evaluation of Explanations} \textbf{Our evaluation of the model's interpretability is indirect}. We use performance on the laboratory abnormality prediction task as a proxy for the quality of the explanation, demonstrating that the ECG embedding contains physiologically relevant information. However, this assessment \textbf{does not establish a causal relationship} between the predicted abnormalities and the final diagnoses, nor does it explicitly highlight which specific abnormalities are most salient for a given diagnostic prediction from the clinician's perspective. The model learns strong correlations, but further investigation, potentially involving causal inference methods or direct clinician feedback studies, would be required to untangle these into clinically actionable insights.

% \paragraph{multimodal learning} when we chose the joint-embedding loss function we select the ones that could be later generalized to a other modes then just the tabular data, in especial the text data. the mimic-iv also has text information with rich information that arguably could improve even more the classification task. also a text-like prediction as explanation is even better than the lab abnormalities one. we plan to incorporate the text in a future work generalizing the joint-embedding loss function to align the embedding of all modes to the signal one. we think keeping one latent representation as an refenrence, or anchor, scale better than the sum of all pair combinations (the naive approach).
\paragraph{Expanding Multimodal Learning to Text} Our choice of a joint-embedding framework was deliberate, with future extensions in mind—in particular, the generalization of a $n>2$ multimodal training with the integration of unstructured text data, such as clinical notes, which are available in MIMIC-IV dataset. This rich textual information could further improve classification performance and enable the generation of more natural, text-based explanations for the model's predictions. We plan to generalize the joint-embedding objective to align representations from all three modalities (ECG, tabular, text). Crucially, we propose using the ECG's latent space as a central anchor, aligning the other modalities to it. This \emph{anchor-based} approach, inspired by the teacher-student approach in some SSL methods, is hypothesized to scale more effectively than the naive alternative of optimizing all pairwise combinations of modalities.

% \paragraph{conclusion} conclude the text, highlighting (i) the joint-embedding as a way to distill knowledge from other modes of data while keeping the ability to inference with just one of these; (ii) the other-mode prediction as a better way to explain the classification prediction in comparison to heatmaps in the input space
\paragraph{Conclusion} In this work, we introduced a novel three-stage training paradigm to develop a powerful and practical ECG-based diagnostic model. Our primary contribution is show evidence that a self-supervised, joint-embedding objective can effectively transfer knowledge from multimodal clinical data into a unimodal signal encoder for ECG classification. This allows the model to leverage rich contextual information during training while retaining the practical advantage of requiring only the ECG signal at inference time. Furthermore, we established a new mechanism to aid decision making by training the encoder to predict associated laboratory abnormalities from the same latent representation. This method of \emph{other-mode prediction} provides indirect explanations that are grounded in concrete physiological concepts, offering a more intuitive and potentially more clinically useful alternative to the commonly input-space heatmaps.

\bibliography{papel/refs}
\bibliographystyle{plainnat}

%%%%%%%%%%%%%%%%%%%%%%%%%%%%%%%%%%%%%%%%%%%%%%%%%%%%%%%%%%%%
\newpage
\appendix
\section*{Appendix}

% \section{Detailed Classification Analysis}
% here we have the metrics for each diagnoses leveraging \{code chapters, condition prevalence, etc\}

% \section{Detailed Estimation Analysis}
% same thing but for lab abnormalities
\section{Related Work}
Our work is primarily related to prior work on eXplainable Artificial Intelligence (XAI), self-supervised learning, and multimodal training in healthcare. To the best of our knowledge, our work is the first attempt to unify insights from these areas for automatic ECG classification.

\paragraph{eXplainable Artificial Intelligence in Healthcare} The growing adoption of deep learning models in medical applications, such as the automatic classification of electrocardiograms (ECG), has demonstrated impressive performance in identifying various cardiac abnormalities \citep{liu2021deep,petmezas2022state}. However, the black box nature of many of these deep learning algorithms limits their widespread acceptance by healthcare professionals \citep{reyes2020interpretability,adeniran2024explainable,rosenbacke2024explainable,koccak2025bias}. The lack of transparency in the decision-making processes of these models generates reluctance, especially in high-risk environments. Traditional XAI methods, such as activation maps in the input space, are often employed to visualize the regions of the input that most influence a model's prediction \citep{chaddad2023survey}. However, in the medical context, these visual explanations are often insufficient. Cardiologists interpret ECGs based on established clinical concepts, such as P-wave morphology or ST-segment elevation, not on raw signal intensity. Explanations that do not resonate with this established clinical lexicon are difficult to interpret and, therefore, less useful for diagnostic validation or learning \citep{borys2023explainable,zhang2023revisiting}. In response to these limitations, there is a growing demand for more intuitive and clinically relevant XAI approaches that can bridge the gap between abstract AI predictions and human understanding, thereby fostering greater trust and facilitating the integration of AI into clinical decision support systems \citep{hulsen2023explainable,gupta2024comparative}.

% \paragraph{self-supervised learning} discuss briefly the most famous self-supervised learning methods highlighting the difference between: joint-embedding -> align the embeddings from augmented views, reference at least barlow-twins \citep{zbontar2021barlow}, vicreg \citep{bardes2021vicreg}, dino \citep{caron2021emerging}, simclr \citep{chen2020simple}; and, reconstruction based methods -> reconstruct the input from an augmented (distorced) view, reference at least vae \citep{kingma2013auto}, gan \citep{goodfellow2020generative}, diffusion \citep{ho2020denoising}, bert \citep{devlin2019bert}
\paragraph{Self-Supervised Learning} Self-supervised learning (SSL) has emerged as a powerful paradigm for learning meaningful representations from unlabeled data. Methodologies can be broadly categorized into two families. The first, joint-embedding methods, learn by enforcing consistency between the embeddings of two or more augmented views of the same data point. These approaches, such as SimCLR \citep{chen2020simple}, DINO \citep{caron2021emerging}, Barlow Twins \citep{zbontar2021barlow}, and VICReg \citep{bardes2021vicreg}, primarily use a discriminative objective to pull representations of the same instance together while preventing representational collapse. The second family consists of reconstruction-based methods, which learn by reconstructing the original input from a corrupted or masked version. This generative approach includes classic models like Variational Autoencoders (VAEs) \citep{kingma2013auto} and Generative Adversarial Networks (GANs) \citep{goodfellow2020generative}, as well as modern powerhouses like Denoising Diffusion models \citep{ho2020denoising} and Masked Language Models like BERT \citep{devlin2019bert}. Our pre-training stage falls into the joint-embedding category, leveraging its strength in learning abstract, transferable features.

% \paragraph{Multimodal with text supervision} Multimodal learning is an advanced artificial intelligence paradigm that integrates information from various data sources \citep{radford2021learning}, such as medical images, physiological signals, and clinical text, to achieve a more comprehensive understanding and improve diagnostic accuracy. In healthcare, this approach is particularly valuable because patient conditions are often characterized by heterogeneous data. Recent advances in Large Language Models (LLMs) further highlight the potential of this field, demonstrating strong capabilities in integration, generalization, and reasoning across diverse medical data modalities, offering promising avenues for generating interpretable disease diagnosis reports \citep{liu2024zero,jin2023medcpt,xiao2023integrating,zhang2024large}.
% \paragraph{multimodal training in healthcare} here we need to change the focus on text to a broader multimodal training
\paragraph{Multimodal Training in Healthcare} Clinical reality is inherently multimodal; a patient's status is defined by a combination of physiological signals, lab values, imaging data, and clinical notes. Multimodal machine learning aims to integrate these heterogeneous data sources to build more robust and accurate models for a holistic understanding of patient health \citep{kline2022multimodal,lin2024has}. A common strategy is feature fusion, where representations from different modalities are combined at an early, intermediate, or late stage to make a final prediction \citep{xiao2023integrating}. For instance, a late-fusion model might combine the outputs of separate encoders for ECGs and tabular data, as is done in our multimodal baseline \citep{alcaraz2025enhancing}. While powerful, these fusion-based approaches typically require all data modalities to be present at inference time, which can be a significant practical barrier in clinical workflows. Our work presents an alternative: using multimodal data during training via a knowledge transfer objective to enrich a unimodal encoder, thereby gaining the benefits of multimodal context without the constraint of multimodal input during deployment.

% \paragraph{Automatic ECG classification} just describe using deep learning for the classification task of ecgs, reference at least the baselines strodthoff2024prospects \citep{strodthoff2024prospects}, alcaraz2024cardiolab \citep{alcaraz2024cardiolab}, alcaraz2025enhancing \citep{alcaraz2025enhancing}
\paragraph{Automatic ECG Classification} Deep learning has become the state-of-the-art approach for the automatic interpretation of ECGs, capable of identifying a wide range of cardiac and even non-cardiac conditions with high accuracy. Our work builds directly upon recent advancements in this area. We situate our contribution relative to three distinct but related lines of research: (i) supervised, unimodal models that classify diagnoses from the ECG signal alone \citep{strodthoff2024prospects}; (ii) multimodal models designed to predict laboratory abnormalities from ECG and routine clinical data \citep{alcaraz2024cardiolab}; and (iii) state-of-the-art multimodal fusion models that achieve high performance but require all data sources at inference time, serving as a practical upper bound for the classification task \citep{alcaraz2025enhancing}.

% \section{Three-stage algorithm}
% We organised the approach into a fluxogram, as despicted in Figure~\ref{fig:architecture}.

% \begin{figure}[ht]
%   \centering
%   \centerline{\includegraphics[width=\columnwidth]{papel/fig/architecture.png}}
%   \caption{Schematic of the proposed multimodal training architecture. Our method begins with a (i) joint-embedding pre-training stage, where an ECG encoder $\Phi_x$ learns to produce representations $H_x$ that are aligned with embeddings from tabular clinical data $M$. Subsequently, this single, enriched encoder is finetuned for two downstream tasks: (ii) a primary multi-label diagnosis classification task $\mathcal{L}_{c}$; and, (iii) a secondary laboratory abnormality reconstruction-like task $\mathcal{L}_{r}$, which provides the mechanism for model interpretability. Crucially, only the ECG signal $X$ is required at inference time.}
%   \label{fig:architecture}
% \end{figure}

\section{Detailed Experiments}
We evaluate our proposed joint-embedding and reconstruction learning paradigm on two distinct downstream tasks: multi-label diagnosis classification; and, laboratory abnormality prediction. To conduct this evaluation, our study utilizes the MIMIC-IV-ECG dataset, a large, publicly available resource uniquely suited for our multimodal research. It contains over 200,000 12-lead electrocardiograms from patients admitted to the Beth Israel Deaconess Medical Center, with a critical linkage to the rich clinical data within the main MIMIC-IV database \citep{johnson2023mimic}. This linkage provides access to a comprehensive set of corresponding tabular data for each ECG, including: (i) laboratory test results; (ii) vital signs and biometrics; and, (iii) ICD-coded discharge diagnoses. The availability of synchronously recorded signals and comprehensive clinical context is fundamental to our approach, as it enables the self-supervised pre-training and provides the ground-truth labels for our downstream tasks. Following established prior work, our experiments focus on the subset of data originating from the emergency department to reflect a challenging and practical clinical screening scenario \citep{strodthoff2024prospects}. Our experimental setup, including data splits, backbone architecture, and evaluation algorithms, rigorously follows the protocol established in \citet{alcaraz2025enhancing} to ensure a fair comparison with all baselines. For the purpose of the lab prediction task, we discriminate the tabular data into two types: the valuable laboratory exam values; and the remaining tabular features, which we term routine clinical data (biometrics and vital signs). All results are reported as the macro-averaged Area Under the Receiver Operating Curve (AUROC) on the held-out test set.

\end{document}